\pdfoutput=1

\documentclass[11pt]{article}

\usepackage[]{naacl2021}

\usepackage{times}
\usepackage{latexsym}

\usepackage[T1]{fontenc}

\usepackage[utf8]{inputenc}

\usepackage{microtype}
\usepackage{graphicx}
\usepackage{mathrsfs}
\usepackage{dutchcal}
\usepackage{amsfonts}
\usepackage{amsmath}
\usepackage{bm}
\usepackage{multirow}
\usepackage{booktabs}

\usepackage{caption}
\usepackage{subcaption}

\usepackage{siunitx}
\usepackage{pgfplots}
\usepackage{tikz-dependency}
\pgfplotsset{every axis/.append style={
                    axis x line=middle,    
                    axis y line=middle,    
                    axis line style={->}, 
                    ylabel near ticks,
                    xlabel near ticks,
                    xlabel={$x$},          
                    ylabel={$y$},          
                    label style={font=\tiny},
                    tick label style={font=\tiny},
                    legend style={font=\tiny},
                    legend pos=outer north east
                    }}

\tikzset{>=stealth}

\title{Improving Pretrained Models for Zero-shot Multi-label Text Classification through Reinforced Label Hierarchy Reasoning}

\author{
    Hui Liu$^1$ \qquad
    Danqing Zhang$^2$ \qquad 
    Bing Yin$^2$ \qquad
    Xiaodan Zhu$^1$ \\
    $^1$Ingenuity Labs Research Institute \& ECE, Queen’s University, Canada \\ 
    {\tt \{hui.liu, xiaodan.zhu\}@queensu.ca } \\
    $^2$Amazon.com Inc, Palo Alto, CA, USA\\
    {\tt \{danqinz, alexbyin\}@amazon.com }
}

\begin{document}
\maketitle
\begin{abstract}


Exploiting label hierarchies has become a promising approach to tackling the zero-shot multi-label text classification (ZS-MTC) problem.
Conventional methods aim to learn a matching model between text and labels, using a graph encoder to incorporate label hierarchies to obtain effective label representations  \cite{rios2018few}.
More recently, pretrained models like BERT \cite{devlin2018bert} have been used to convert classification tasks into a textual entailment task \cite{yin-etal-2019-benchmarking}. 
This approach is naturally suitable for the ZS-MTC task.
However, pretrained models are underexplored in the existing work because they do not generate individual vector representations for text or labels, making it unintuitive to combine them with conventional graph encoding methods.
In this paper, we explore to improve pretrained models with label hierarchies on the ZS-MTC task.
We propose a Reinforced Label Hierarchy Reasoning (RLHR) approach to encourage interdependence among labels in the hierarchies during training.
Meanwhile, to overcome the weakness of flat predictions, we design a rollback algorithm that can remove logical errors from predictions during inference.
Experimental results on three real-life datasets show that our approach achieves better performance and outperforms previous non-pretrained methods on the ZS-MTC task.

\end{abstract}

\section{Introduction}

\begin{figure}[t]
  \includegraphics[width=\linewidth]{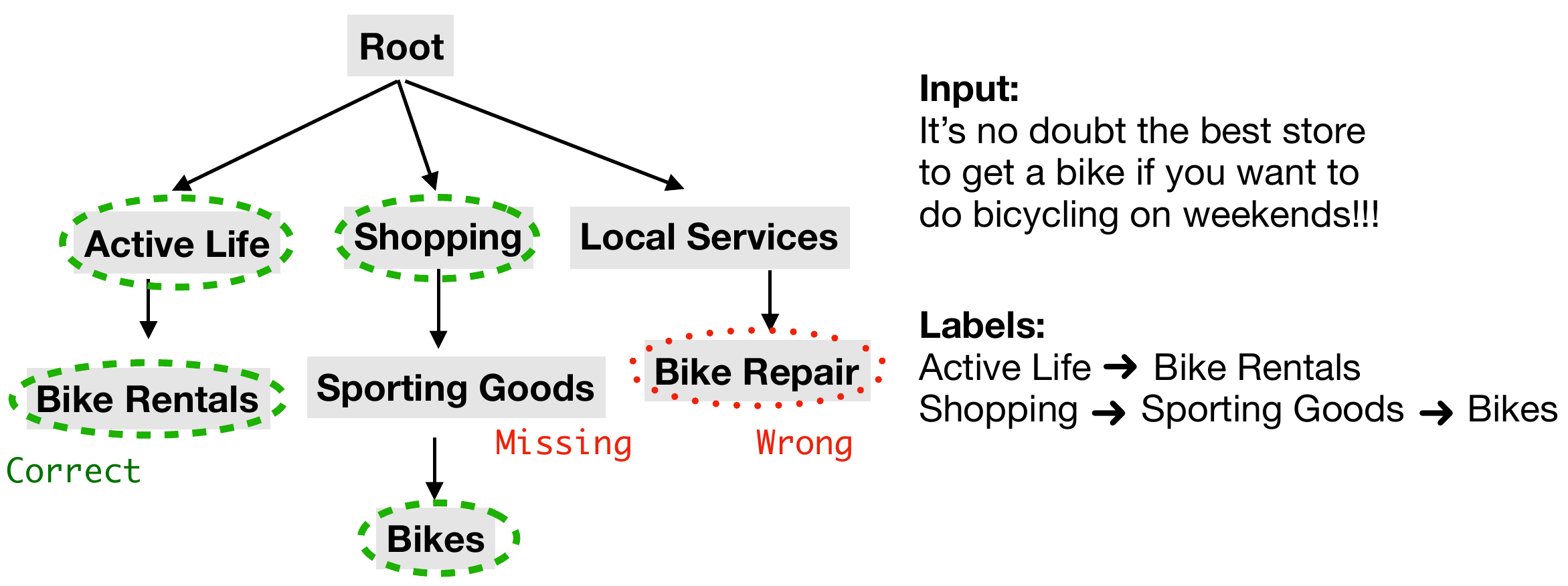}
  \caption{An example of label hierarchy and predictions with logical errors.
            Circled labels are model predictions without incorporating label hierarchy.}
  \label{fig:hier_example}
  \vspace{-15pt}
\end{figure}

Multi-label text classification (MTC) is a basic NLP problem that underlies many real-life applications like product categorization \cite{partalas2015lshtc} and medical records coding \cite{du2019ml}.
The labels in the output space are  often interdependent and in many applications organized in a hierarchy, as shown in the example in Figure \ref{fig:hier_example}. 
A significant challenge for real-life development of MTC applications is severe deficiencies of annotated data for each label in the hierarchy, which demands better solutions for zero-shot learning.
The existing zero-shot learning for multi-label text classification (ZS-MTC) mostly learns a matching model between the feature space of text and the label space \cite{ye2020zero}.
In order to learn effective representations for labels, a majority of existing work incorporates label hierarchies via a label encoder designed as Graph Neural Networks (GNNs) that can aggregate the neighboring information for labels \cite{chalkidis2020empirical,lu2020multi}.

Recently, pretrained models like BERT \cite{devlin2018bert} have been widely used as strong matching models due to their superior representation ability \cite{qiao2019understanding}. 
They have been applied to convert a classification task to a textual entailment task, by treating the text to be classified as the premise, and its label as the hypothesis, which is naturally suitable for the ZS-MTC study \cite{yin-etal-2019-benchmarking}.
However, the problem of this approach is that pretrained models cannot generate individual vector representations for labels---a label is coupled with the corresponding text in learning joint representation---thus conventional methods, like GNNs which utilize the label hierarchy to obtain better label representations, cannot be directly applied to pretrained models, making them underexplored in the existing research.


Although pretrained models have shown potential on ZS-MTC, as discussed above, it is not intuitive to introduce structural information of label hierarchies to the learning procedure.
Flattening all the labels without considering their hierarchical structures, however, will result in predictions that contain logical errors, which are known as the class-membership inconsistency \cite{silla2011survey}.
The problem will be even more salient for pretrained models because they only take the literal tokens of the labels as input.
An example with logical errors is shown in Figure \ref{fig:hier_example}.
Without label hierarchy information, the model correctly predicts \textit{Bikes} as a true label, but fails to predict its parent label, \textit{Sporting Goods}.
Meanwhile, the model does not choose the label \textit{Local Services} while predicting its child label \textit{Bike Repair} due to the fact that \textit{Bike Repair} has tokens similar to those in the input text. 

To overcome the forementioned weakness, we propose a Reinforced Label Hierarchy Reasoning (RLHR) approach to introduce label structure information to pretrained models. 
Instead of regarding labels to be independent, we cast ZS-MTC as a deterministic Markov Decision Process (MDP) over the label hierarchy.
An agent starts from the root label and learns to navigate to the potential labels by hierarchical deduction in the label hierarchy.
The reward is based on the correctness of the deduction paths, not simply on the correctness of each label.
Thus the reward received by one predicted label will be determined by both the label itself and other labels on the same path, which can help to strengthen the interconnections among labels.
Meanwhile, we find that the hierarchical inference method \cite{huang2019hierarchical} will broadcast the errors arising at the higher levels of label hierarchies.
Thus we further design a rollback algorithm based on the predicted matching scores of labels to reduce the logical errors in the flat prediction mode during inference.
We apply our approach to different pretrained models and conduct experiments on three real-life datasets.
Results demonstrate that pretrained models outperform conventional non-pretrained methods by a substantial margin.
After being combined with our approach, pretrained models can attain further improvement on both the classification metrics and logical error metrics\footnote{Code and data available at \url{https://github.com/layneins/Zero-shot-RLHR}}.
We summarize our contributions as follows:
\begin{itemize}
    \itemsep 0em
    \item We demonstrate that pretrained models outperform conventional methods on ZS-MTC.
    \item We design a novel Reinforced Label Hierarchy Reasoning (RLHR) approach and a matching-score-based rollback algorithm to introduce the structural information of label hierarchies to pretrained models in both the training and inference stage.
    \item Experiments with different pretrained models are performed on three real-life datasets. We show the effectiveness of our proposed approach and provide detailed analyses.
\end{itemize}

\section{Related Work}

Exploiting the prior distribution of the label space has proven to be an effective method to tackle the multi-label text classification problem because it can provide the model with information about the label structure.
\citet{mao2019hierarchical,huang2019hierarchical} took the explicitly represented label hierarchy as the structural information, while \citet{wu2019learning} assumed the prior distribution to be implicit and trained their model to learn the distribution during learning.

Leveraging the label hierarchy to tackle ZS-MTC has shown to be promising in previous work, which mostly aimed to learn a matching model between texts and labels.
\citet{chalkidis2020empirical,chalkidis-etal-2019-large,xie2019ehr} adopted Label-Wise Attention Networks to encourage interactions between text and labels.
\citet{rios2018few,lu2020multi} used Graph Neural Networks to capture the structural information in the label hierarchy.
However, few existing works investigate the effectiveness of pretrained models on the ZS-MTC task, despite pretrained models being effective as matching models for many natural language processing tasks \cite{ma2019universal,qiao2019understanding,nogueira2019multi}.

The logical error problem in flat predictions has been widely discussed in previous MTC work \cite{silla2011survey,wehrmann2018hierarchical,mao2019hierarchical}, which is mostly solved through a hierarchical procedure during inference.
In our work, we will investigate such a method and see that the hierarchical inference method is not optimal for pretrained models on the ZS-MTC task because it broadcasts errors top-down in the label hierarchy.

Path reasoning is effective for exploiting explicit relationships in structured data, which can be combined with reinforcement learning, e.g., knowledge graph reasoning \cite{wan2020reasoning,xian2019reinforcement,xiong-etal-2017-deeppath}. 
We propose to introduce the label hierarchy to pretrained models through path reasoning, with the aim to strengthen the interconnections between labels. 
To the best of our knowledge, our work is the first to improve pretrained models through label hierarchies for ZS-MTC.

\section{Problem Formulation}

\subsection{Label Hierarchy Reasoning}

In general, a label hierarchy is defined as $\mathcal{G}=(\mathcal{L}, \mathcal{E})$, where $\mathcal{L}$ and $\mathcal{E}$ are a set of labels and relations, respectively. The latter represent parent-child relations between labels.
The root of $\mathbcal{G}$ is a special label $\mathbb{R}$. 
A data instance $\mathbcal{x}$ is defined as a tuple $(T, P)$ with $T$ as the input text and $P=\{p_1, p_2, \cdots, p_N\}$ as deduction paths, and a path $p_i=\{\mathbb{R}, l_i^1, \cdots, l_i^{K-1}, l_i^K\}$ where $l_i^k \in \mathcal{L}$ is at the $k^{th}$ layer of $\mathcal{G}$ and $l_i^{k-1}$ is the parent of $l_i^k$.
A deduction path must be contiguous, starting with $\mathbb{R}$, and is not required to terminate at a leaf label.

\subsection{Zero-shot Multi-label Text Classification}

Let $\mathcal{L}_s$ and $\mathcal{L}_u$ denote the seen and unseen labels, respectively, where $\mathcal{L}_s \cup \mathcal{L}_u=\mathcal{L}$.
Given a training set $\mathcal{D}^s=\{\mathbcal{x}^s_i\}_{i=1}^{N_1}$ where the labels of $\mathbcal{x}^s_i$ are all seen labels, we aim to learn a matching model $f(\mathcal{D}^s; \theta)$ and make prediction on $\mathcal{D}^u=\{\mathbcal{x}^u_i\}_{i=1}^{N_2}$.
Some deduction paths of $\mathbcal{x}^u_i$ consist of seen labels while some contain both seen and unseen labels.
Notice that the children of an unseen label are also unseen labels. 
Evaluations on $\mathcal{D}^u$ will be conducted in two settings: (1) evaluate the performance on $\mathcal{L}_u$, which is known as the zero-shot (ZS) setting, and (2) evaluate the performance on $\mathcal{L}_s \cup \mathcal{L}_u$, which is the generalized zero-shot (GZS) setting \cite{huynh2020shared}.

\section{Methodology}

The goal of our RLHR approach is to learn a policy $\mathcal{P}$ that can make more consistent predictions by traversing the label hierarchy $\mathcal{G}$ to generate deduction paths.
Given a training instance $\mathcal{x}$, an agent will start from the root $\mathbb{R}$ and follow $\mathcal{P}$ at each time step to extend the deduction paths by navigating to the children labels at the next level.
By measuring the correctness of the generated deduction paths with reinforcement learning (RL), the label hierarchy is introduced to the model during the training time and the interconnections of labels will hence be strengthened, which can help to reduce logical errors in prediction. 
As we will show in our experiments, hierarchical inference, which is used in previous work \cite{mao2019hierarchical}, will propagate the errors occurring at the high levels of hierarchies during inference, resulting in inferior performance. 
Thus we still adopt the flat prediction during inference, but further design a rollback algorithm based on the structure of $\mathcal{G}$ and the predicted matching scores. 
We will introduce the details of our proposed RLHR and the rollback algorithm in the following subsections.

\begin{figure*}[t]
  \centering
  \includegraphics[width=0.85\linewidth]{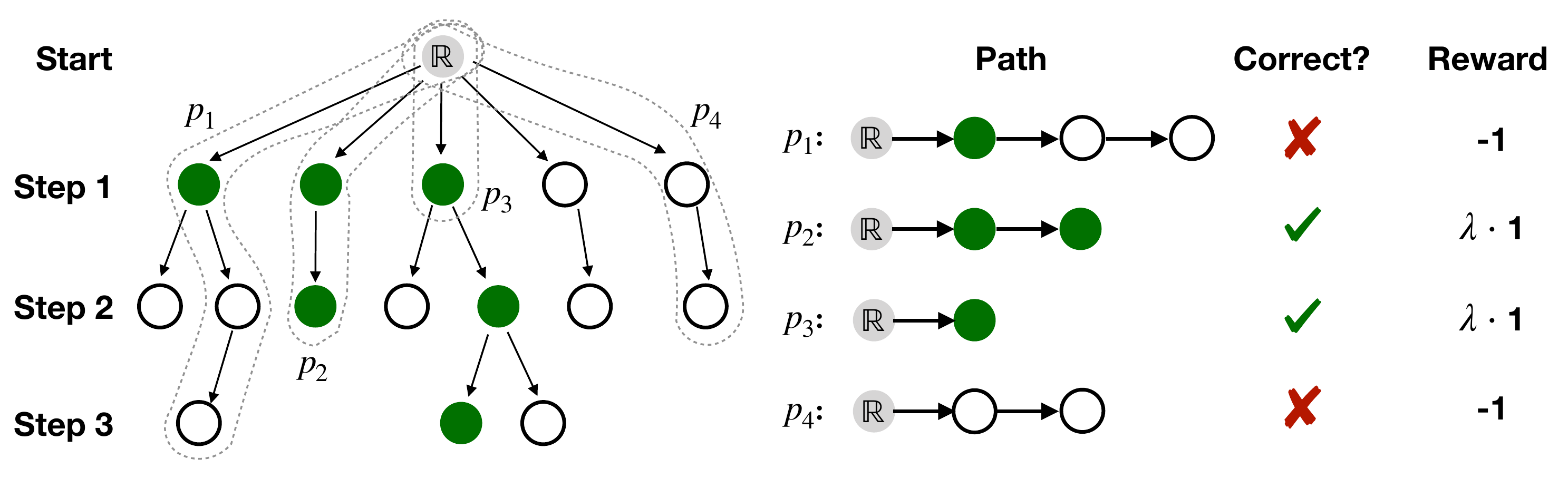}
  \caption{An example of our RLHR approach with $M=4$.
            Green circles are the ground truth labels.
            $p_1, p_2, p_3$, and $p_4$ are four sampled deduction paths, where $p_3$ ends before it arrives at a leaf label.
            }
  \label{fig:rl_example}
  \vspace{-10pt}
\end{figure*}

\subsection{Base Model}

Our base model adopts pretrained models $\mathcal{M}$, e.g., BERT \cite{devlin2018bert}, which have proven to be effective in matching modelling.
Given the input text $T$ and the label $l$, we follow \citet{yin-etal-2019-benchmarking} by transforming the text-label pair into textual entailment representation as ``[CLS] $T$ [SEP] hypothesis of $l$''.
The hidden vector $v_{cls}$ of [CLS] is regarded as the aggregate representation and will be used in the classification layer to calculate the matching score $ms$.
The overall calculation process of $ms$ is abbreviated as:
\begin{gather}
    ms = \mathcal{M}(T, l)
\end{gather}
If $ms \geq \gamma$ where $\gamma$ is a threshold, we then say $T$ belongs to label $l$. In experiments $\gamma$ is set to be 0.5.

\subsection{Reinforced Label Hierarchy Reasoning (RLHR)}

Different from vanilla pretrained models that rely on flat prediction during training, we propose to formulate the ZS-MTC task as a deterministic Markov Decision Process (MDP) over label hierarchies.
For the input text, the agent trained by RLHR will predict $M$ deduction paths from the root label $\mathbb{R}$.
When all deduction paths are generated, the rewards will be received, which are determined by the correctness of the paths.
An overall illustration of the RLHR approach is shown in Figure \ref{fig:rl_example}.
We introduce the details of the RL modules in this subsection.

\subsubsection{States}
Maintaining just one deduction path for one data instance will result in an inefficient learning process. However, the number of potential deduction paths will increase exponentially as the model goes deeper into the lower levels of the hierarchies. To maintain a good trade-off between computational resources and time efficiency, we keep the beam of deduction paths to be $M$.
Thus for a data instance $\mathcal{x}$, the global state $S^k$ at step $k$ is composed of the sub-states of $M$ deduction paths:
\begin{gather}
    S^k = \{s_i^k\}_{i=1}^M
\end{gather}
The sub-state $s_i^k$ for deduction path $p_i$ at step $k$ is defined as a tuple $(T, l_i^k)$, where $T$ is input text and $l_i^k$ is the label. 

\subsubsection{Actions}

The complete action space $A_i^k$ of sub-state $s_i^k$ is defined as all possible child labels of label $l_i^k$:
\begin{gather}
    A_i^k = \{l| l \in C(l_i^k)\}
\end{gather}
where $C(l_i^k)$ denotes the child labels of $l_i^k$.
For the deduction path $p_i$ at the time step $k$, an action $a_i^k$ is to select one label $l_i^{k+1}$ from $A_i^k$. 
Notice that the agent may not select any labels from $A_i^k$, which means path $p_i$ ends before it arrives at a leaf label and a ``stop'' action is taken.
By adding this ``early stop'' mechanism, we can make the agent automatically learn when to stop assigning new labels to the deduction paths.

\subsubsection{Policy}

We parameterize the action $a_i^k$ by a policy network $\pi(\bm{\cdot}|s, A; \theta)$ where $\theta$ is parameters.
For deduction path $p_i$ at time step $k$, the policy network takes as input the state $s_i^k$ and the corresponding action space $A_i^k$, emitting the matching score of each action in $A_i^k$, which is calculated by the base pretrained model $\mathcal{M}$. 
Finally an action $a_k$ is sampled based on the matching score distribution of the actions in $A_i^k$.
The calculation is formulated as follows:
\begin{gather}
    \pi(\bm{a_i^k}|s_i^k, A_i^k; \theta) = \{\mathcal{M}(T, l) | l \in A_i^k\}\\
    a_i^k \sim \pi(\bm{a_i^k}|s_i^k, A_i^k; \theta)
\end{gather}

\subsubsection{Reward}

In our approach, the reward is based on the correctness of a complete deduction path.
Instead of treating all labels to be flat, our approach encourages the interdependence among the labels.
The reward received by a label $l_i^k$ is not only decided by the correctness of itself but also the correctness of other labels on the same deduction path $p_i$.
Given the golden deduction paths $\hat{P}=\{\hat{p}_1, \hat{p}_2, \cdots, \hat{p}_N\}$, $p_i$ will obtain a positive reward if $p_i$ is in $\hat{P}$ or $p_i$ is a sub-path of a path in $\hat{P}$.
Formally the reward of path $p_i$ is defined as:
\begin{equation}\label{eq:reward}
r_i = 
\begin{cases}
\lambda \cdot 1, & \text{~if~} p_i \subseteq \hat{p}_j \text{~where~} \hat{p}_j \in \hat{P}  \\
-1, & \text{~otherwise,}
\end{cases}
\end{equation}
where $\lambda$ is a hyper-parameter for scaling.
Under most circumstances, the number of wrong deduction paths will be greater than the correct ones. 
The problem will be even more severe for the MTC tasks because the distribution of positive labels and negative labels is usually imbalanced given a data instance $\mathcal{x}$.
A larger $\lambda$ can encourage the model to focus more on the correct paths.

Notice that our approach differs from existing methods which adopt hierarchical classification~\cite{sun2001hierarchical,peng2018large}.
A hierarchical classification method based on the label hierarchy can only cast the influence from parent label to child label, while in our approach the influence is mutual between parent label and child label, which can hence strengthen the reasoning ability of the models.

\subsubsection{Optimization}

Our goal is to learn a stochastic policy $\pi$ that maximize the expected total reward $J(\theta)$ of the M sampled deduction paths, which can be formulated as:
\begin{gather}
    J(\theta) = E_{\pi(a|s)}[\sum_{i=1}^M r_i(\bm{s}, \bm{a})]
\end{gather}
where $\theta$ is the parameter of policy network.
We adopt policy gradient \cite{sutton2000policy} as the optimization algorithm which updates $\theta$ as:
\begin{gather}
    \theta \leftarrow \theta + \eta \nabla_{\theta} \tilde{J}(\theta)
\end{gather}
where $\eta$ is the discount learning rate.
Since there are multiple deduction paths for one data instance, the gradient can be approximated by
\begin{gather}
    \tilde{J}(\theta) = \frac{1}{M} \sum_{i=1}^M \sum_k \log \pi(a_i^k|s_i^k; \theta) \cdot (r_i - r_b)
\end{gather}
$r_b$ is a constant for the stabilization of the training procedure, for which we use the average reward of the last training epoch in our experiments.

\subsection{Inference Rollback}

Existing methods mostly adopt the hierarchical inference method \cite{mao2019hierarchical}, which will avoid logical errors, i.e., class-membership inconsistency \cite{silla2011survey}, but bring a serious problem: the prediction errors made at the high levels of a hierarchy are often severely propagated to the lower levels.
For instance, if a correct label at the first layer is missing, then all the descendant labels will not be considered during inference.
This will no doubt harm the performance.
On the contrary, if the model still makes flat prediction, all labels will be visited during inference, while more logical errors will probably arise. 

To overcome the forementioned weaknesses, we propose a rollback algorithm during the inference stage based on the predicted matching scores of all labels.
For a data instance $\mathcal{x}$, we obtain the predicted labels in flat prediction mode as $P$, which consists of two parts: (1) labels that can form complete deduction paths, and (2) labels with logical errors, which we denote as $P_e=\{l_1^{k_1}, l_2^{k_2}, \cdots, l_N^{k_N}\}$.
For a label $l_i^{k_i} \in P_e$, we extract its deduction path from $\mathcal{G}$ as $p_i=\{\mathbb{R}, l_i^1, \cdots, l_i^{k_i-1}, l_i^{k_i}\}$ and their corresponding predicted matching scores $\{1, ms_i^1, \cdots, ms_i^{k_i-1}, ms_i^{k_i}\}$\footnote{Root label $\mathbb{R}$ always has a matching score 1.}.
Meanwhile we set a rollback threshold $\mu^k$ for the labels in the $k^{th}$ layer of $\mathcal{G}$, where $\{\mu_k\}$ are hyper-parameters tuned on the development set.
As long as the matching scores meet the requirements 
$$\{ms_i^j \geq \mu^j\}_{j=1}^{k_i-1},$$ we add the labels in $p_i$ back to $P$.
Otherwise label $l_i^{k_i}$ will be removed from $P$.

The motivation behind this matching-score-based rollback algorithm is that for a label hierarchy $\mathcal{G}$, the labels at higher-level hierarchy contain more training instances but their meaning are more abstract, while the labels at lower levels are more specific such as the labels ``Active Life'' and ``Bike Rentals'' in Figure \ref{fig:hier_example}.
Pretrained models just take as input the literal tokens of a label and thus are possible to obtain a better performance on certain labels at the lower levels than those at higher levels.

\section{Experiments}

\subsection{Experimental Setup}

\subsubsection{Datasets}

We conduct experiments on three real-life datasets from different domains; the details are provided in Table \ref{tab:data_statis}.
Yelp\footnote{https://www.yelp.com/dataset} is a customer review dataset, in which we need to classify customer reviews into  correct business categories.
WOS \cite{Kowsari2018HDLTex} is a scientific paper dataset which provides the abstracts of published papers and the corresponding topics. QCD is a query classification dataset we create for the ZS-MTC task.
It is composed of search queries and target product types, which is collected from e-commerce websites. 
The layer numbers of the label hierarchies in Yelp, WOS and QCD are 4, 2, and 3, respectively. 
For examples of the three datasets, please refer to Appendix \ref{sec:data_example}.

\begin{table}[t]
\centering
\setlength{\tabcolsep}{1.5pt}
\begin{tabular}{c|cccc|cc}  
\toprule
    \multirow{2}{*}{Dataset} & \multicolumn{4}{c|}{Docs} & \multicolumn{2}{c}{Labels} \\
        & \#Train & \#Dev & \#Test & Avg($\lvert L \rvert $) & seen & unseen \\
    \midrule
    Yelp    & 187153  & 10858   & 10858   & 3.80 & 466    & 71 \\
    WOS     & 36397   & 5294    & 5294    & 2.00 & 122    & 28 \\
    QCD     &  177423 & 12277   & 12277   & 4.69 & 243    & 93 \\
\bottomrule
\end{tabular}
\caption{Dataset Statistics. Avg($\lvert L \rvert $) denotes the average number of labels in one data instance.}
\label{tab:data_statis}
\end{table}

\begin{table*}[t]
\centering
\setlength{\tabcolsep}{2.0pt}
\begin{tabular}{c|c|ccc|c|ccc|c|ccc|c}  
\toprule
    \multirow{2}{*}{Method} & \multirow{2}{*}{Setting} & \multicolumn{4}{c|}{Yelp} & \multicolumn{4}{c|}{WOS} & \multicolumn{4}{c}{QCD} \\
    \cline{3-14}
    &   &  Ma-F & Mi-F & EBF & Err$\downarrow$ & Ma-F & Mi-F & EBF & Err$\downarrow$ & Ma-F & Mi-F & EBF & Err$\downarrow$ \\
    \midrule
    CNN & ZS  & 
    0.33 & 2.02 & \multirow{2}{*}{16.35} & \multirow{2}{*}{0.3211} & 
    0.36 & 4.43 & \multirow{2}{*}{28.22} & \multirow{2}{*}{0.2977} & 
    5.02 & 6.58 & \multirow{2}{*}{26.94} & \multirow{2}{*}{2.9386} \\
        & GZS & 
    1.31 & 14.97 &                      &                      & 
    7.00 & 29.58 &                      &                      & 
    9.66 & 26.22 &                      &                      \\
    \midrule
    CNN & ZS  & 
    4.24 & 7.15 & \multirow{2}{*}{19.38} & \multirow{2}{*}{0.9303} & 
    0.54 & 4.53 & \multirow{2}{*}{26.88} & \multirow{2}{*}{0.3079} & 
    5.02 & 7.09 & \multirow{2}{*}{28.24} & \multirow{2}{*}{4.3923} \\
    +LWAN  & GZS & 
    4.67 & 19.26 &                      &                      & 
    6.81 & 29.00 &                      &                      & 
    10.03 & 28.86 &                      &                      \\
    \midrule
    \multirow{2}{*}{ZAGCNN} & ZS  & 
    17.94 & 18.75 & \multirow{2}{*}{28.24} & \multirow{2}{*}{1.3136} &
    12.02 & 17.17 & \multirow{2}{*}{24.72} & \multirow{2}{*}{2.5827} & 
    5.22 & 10.01 & \multirow{2}{*}{\textbf{40.65}} & \multirow{2}{*}{2.0212} \\
        & GZS & 
    16.30 & 25.97 &                      &                      & 
    19.59 & 36.37 &                      &                      & 
    23.85 & \textbf{42.52} &                      &                      \\
    \midrule
    \multirow{2}{*}{DistilBERT} & ZS  & 
    41.42 & 40.33 & \multirow{2}{*}{30.44} & \multirow{2}{*}{0.4039} & 
    70.69 & 65.19 & \multirow{2}{*}{55.18} & \multirow{2}{*}{0.5178} & 
    23.68 & 24.95 & \multirow{2}{*}{33.57} & \multirow{2}{*}{1.0854} \\
        & GZS & 
    21.29 & 28.18 &                      &                      & 
    68.03 & 63.64 &                      &                      & 
    24.43 & 34.29 &                      &                      \\
    \cline{1-14}
    \multirow{2}{*}{+RLHR} & ZS  & 
    42.16 & 43.87 & \multirow{2}{*}{40.85} & \multirow{2}{*}{0.3347} & 
    74.56 & 72.44 & \multirow{2}{*}{61.06} & \multirow{2}{*}{0.4732} & 
    24.58 & 27.79 & \multirow{2}{*}{37.46} & \multirow{2}{*}{\textbf{0.8389}} \\
        & GZS & 
    26.95 & 40.43 &                      &                      & 
    71.65 & 68.05 &                      &                      & 
    26.10 & 38.37 &                      &                      \\
    \midrule
    \multirow{2}{*}{BERT} & ZS  & 
    44.49 & 42.61 & \multirow{2}{*}{34.59} & \multirow{2}{*}{0.3755} & 
    77.87 & 77.27 & \multirow{2}{*}{56.69} & \multirow{2}{*}{\textbf{0.1983}} & 
    28.18 & 27.45 & \multirow{2}{*}{36.88} & \multirow{2}{*}{1.2497} \\
        & GZS & 
    23.38 & 31.53 &                      &                      & 
    74.69 & 70.56 &                      &                      & 
    27.04 & 37.20 &                      &                      \\
    \cline{1-14}
    \multirow{2}{*}{+RLHR} & ZS  & 
    \textbf{45.46} & \textbf{48.26} & \multirow{2}{*}{\textbf{49.52}} & \multirow{2}{*}{\textbf{0.2952}} & 
    \textbf{78.46} & \textbf{79.19} & \multirow{2}{*}{\textbf{64.43}} & \multirow{2}{*}{0.2488} & 
    \textbf{28.32} & \textbf{28.80} & \multirow{2}{*}{39.99} & \multirow{2}{*}{1.1984} \\
        & GZS & 
    \textbf{32.09} & \textbf{49.75} &                      &                      & 
    \textbf{75.51} & \textbf{72.62} &                      &                      & 
    \textbf{28.67} & 41.08 &                      &                      \\
\bottomrule
\end{tabular}
\caption{Results of different methods on the three datasets under two settings.
        Ma-F, Mi-F, EBF, and Err denote Macro-F1, Micro-F1, Example-based F1, and logical error rate, respectively. ZS and GZS denote the zero-shot and generalized zero-shot setting.
        $\downarrow$ means the lower the better.
        Bold numbers indicate the best results for each metric.
        All the results are acquired under the flat prediction.}
\label{tab:all_res}
\vspace{-5pt}
\end{table*}

\subsubsection{Implementation Details}

We test our proposed approach with two pretrained models, BERT \cite{devlin2018bert} and DistilBERT \cite{sanh2019distilbert}.
For BERT, we use the uncased base version, which is of 12-layer transformer blocks, 768-dimension hidden state, 12 attention heads and 110M parameters in total.
For DistilBERT, it contains 6-layers transformer blocks, 768-dimension hidden state and 12 attention heads, totally 66M parameters.
For training, we use Adam \cite{kingma2014adam} for optimization and learning rate is set to 1e-6.
Meanwhile we adopt early stopping to avoid overfitting on the training data.
$\lambda$ is set to 30 on Yelp, 20 on QCD, and 5 on WOS, which we will discuss more in Section \ref{sec:influence_of_lambda}.
We set $M$ to 5 with DistilBERT and 3 with BERT by trading off between training time and GPU memory usage.

The RL training procedure is unstable and slow if the agent is trained from scratch \cite{silver2016mastering}.
So with both BERT and DistilBERT, we pretrain the policy network in flat prediction mode on the training data with the learning rate of 1e-5.

\subsubsection{Evaluation Metrics}

In our experiments, we use standard metrics Micro-F1 and Macro-F1 to evaluate the classification performance for both the zero-shot and generalized zero-shot setting.
Meanwhile, we also adopt Example-based F1 \cite{peng2016deepmesh} to measure the performance from the instance level, which is different from Micro/Macro-F1 measuring from the label level.
Though some previous works adopted ranking based metrics \cite{rios2018few} for large-scale MTC, they are not appropriate in our settings because the datasets used in this work contain smaller label space.


For logical errors, we report the \textit{logical error rate}, which is defined as the average number of logical errors in one data instance.
We take the number of logical errors in one data instance as the number of labels that cannot form a complete deduction path.

Evaluation is conducted in two settings: (1) evaluate the performance on unseen labels only, which is the zero-shot (ZS) setting, and (2) evaluate the performance on both seen labels and unseen labels, i.e., the generalized zero-shot (GZS) setting \cite{huynh2020shared}.

\subsection{Baselines}

We use two different types of baselines. 
(1) The type of models where label hierarchy is not utilized, and we use \textbf{CNN} and CNN with Label-Wise Attention Networks (\textbf{CNN+LWAN}) \cite{chalkidis-etal-2019-large} in our experiments. 
(2) The type of models where GNNs are utilized to encode the label hierarchy to capture the label structure information. Specifically we use  \textbf{ZAGCNN} proposed by \citet{rios2018few}.


\begin{table*}[t]
\centering
\setlength{\tabcolsep}{3.0pt}
\begin{tabular}{c|c|ccc|ccc|ccc}  
\toprule
    \multirow{2}{*}{Method} & \multirow{2}{*}{Setting} & \multicolumn{3}{c|}{Yelp} & \multicolumn{3}{c|}{WOS} & \multicolumn{3}{c}{QCD} \\
    \cline{3-11}
    &   &  Ma-F & Mi-F & EBF & Ma-F & Mi-F & EBF & Ma-F & Mi-F & EBF \\
    \midrule
    \multirow{2}{*}{BERT} & ZS  & 
    44.49 & 42.61 & \multirow{2}{*}{34.59} & 
    77.87 & 77.27 & \multirow{2}{*}{56.69} & 
    28.18 & 27.45 & \multirow{2}{*}{36.88} \\
        & GZS & 
    23.38 & 31.53 &                      & 
    74.69 & 70.56 &                      & 
    27.04 & 37.20 &                      \\
    \midrule
    BERT & ZS  & 
    45.11 & 43.46 & \multirow{2}{*}{34.79} & 
    73.68 & 74.14 & \multirow{2}{*}{54.52} & 
    26.67 & 31.33 & \multirow{2}{*}{37.76} \\
    +Hie-Infe & GZS & 
    23.58 & 31.72 &                      & 
    71.24 & 69.02 &                      & 
    26.88 & 38.16 &                     \\
    \midrule
    BERT & ZS  & 
    44.46 & 42.57 & \multirow{2}{*}{34.65} & 
    77.87 & 77.27 & \multirow{2}{*}{58.28} & 
    28.15 & 27.57 & \multirow{2}{*}{36.69} \\
    +Rollback  & GZS & 
    23.35 & 31.56 &                      & 
    75.26 & 71.81 &                      & 
    26.95 & 36.89 &                     \\
    \hline\hline
    \multirow{2}{*}{BERT+RLHR} & ZS  & 
    45.46 & 48.26 & \multirow{2}{*}{49.52} & 
    78.46 & 79.19 & \multirow{2}{*}{64.43} & 
    \textbf{28.32} & 28.80 & \multirow{2}{*}{39.99}\\
        & GZS & 
    32.09 & 49.75 &                      & 
    75.51 & 72.62 &                      & 
    \textbf{28.67} & 41.08 &                      \\
    \midrule
    BERT+RLHR & ZS  & 
    39.57 & 42.91 & \multirow{2}{*}{48.53} & 
    65.82 & 67.93 & \multirow{2}{*}{56.34} & 
    25.34 & \textbf{32.46} & \multirow{2}{*}{\textbf{40.97}} \\
    +Hie-Infe  & GZS & 
    31.22 & 49.2 &                      & 
    65.1 & 67.41 &                      & 
    28.06 & \textbf{42.23} &                     \\
    \midrule
    BERT+RLHR & ZS  & 
    \textbf{45.57} & \textbf{48.32} & \multirow{2}{*}{\textbf{50.01}} & 
    \textbf{78.46} & \textbf{79.19} & \multirow{2}{*}{\textbf{69.32}} & 
    28.03 & 29.71 & \multirow{2}{*}{40.13} \\
    +Rollback & GZS & 
    \textbf{32.17} & \textbf{50.18} &                      & 
    \textbf{77.16} & \textbf{77.26} &                      & 
    28.58 & 41.18 &                     \\
\bottomrule
\end{tabular}
\caption{Performance of our matching-score-based rollback algorithm and the comparison to the hierarchical inference method. 
        Ma-F, Mi-F, EBF, and Err denote Macro-F1, Micro-F1, Example-based F1, and logical error rate, respectively. ZS and GZS denote the zero-shot and generalized zero-shot setting.
        Bold numbers indicate the best results for each metric.
        ``BERT+Hie-Infe'' in the last row means BERT with the hierarchical inference method, which is used in previous work \cite{huang2019hierarchical}.}
\label{tab:rollback_res}
\vspace{-5pt}
\end{table*}

\subsection{Results}

Table \ref{tab:all_res} shows the experimental results of the baseline models and our proposed RLHR approach on three real-life datasets in both the zero-shot and generalized zero-shot setting.

\subsubsection{Classification Performance}

As we can see in Table \ref{tab:all_res} that CNN and CNN+LWAN have poor performance under the ZS setting while the performance under GZS setting is better, which suggests CNN and CNN+LWAN cannot provide accurate predictions for unseen labels due to the lack of label structure information.
In contrast, ZAGCNN, which utilizes the label hierarchy, performs better, particularly on unseen labels, which demonstrates the importance of label hierarchy for ZS-MTC.

On the other hand, pretrained models, including DistilBERT and BERT, both outperform conventional non-pretrained methods with substantial improvements on three datasets, though ZAGCNN shows slight advantages on Micro-F1 and Example-based F1 on the QCD dataset under the GZS setting.
When incorporated with RLHR, the performance of pretrained models can be further improved by a relatively large margin.
We notice that the improvement under GZS setting is more significant than in the ZS setting, suggesting that seen labels benefit more from our RLHR than unseen labels.


\subsubsection{Logical Errors}

As shown in Table \ref{tab:all_res}, utilizing label hierarchies does not necessarily reduce the logical error rate for conventional methods, though it can improve the classification performance.
For example, the logical error rate of ZAGCNN is higher than CNN and CNN+LWAN on Yelp and WOS. 
The logical error rate of pretrained models is generally lower than the conventional methods.
However, pretrained models still face the logical error problem though they perform well on the classification metrics.
We can also see that our RLHR can help reduce the logical error rate for DistilBERT and BERT under most circumstances.

Note that better classification performance does not necessarily lead to a lower logical error rate. 
From Table \ref{tab:all_res}, we can see although CNN and CNN+LWAN perform poorly on classification metrics, they achieve a better logical error rate than ZAGCNN and DistilBERT on the WOS dataset.
Similarly, the logical error rate of BERT is higher than DistilBERT on QCD even though BERT has a better classification performance.
Our proposed RLHR approach can improve both the classification performance and logical error performance, which demonstrates the effectiveness of RLHR.

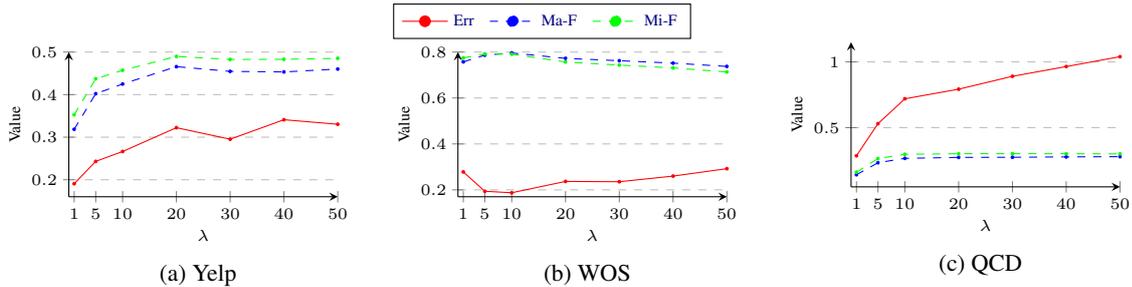
\begin{figure*}[t] \tiny 
  \begin{subfigure}[t]{0.32\textwidth}
  \pgfplotsset{width=\textwidth}
  \ref{bb} \\
\begin{tikzpicture}
    \begin{axis}[
    height=3.5cm,
    xlabel={$\lambda$},
    xtick={1,5,10,20,30,40,50},
  ylabel={Value},
  xmin=0, xmax=50,
        ymin=0.16, ymax=0.50,
    mark size=0.5pt,
    ymajorgrids=true,
    grid style=dashed,
    legend columns=-1,
    legend entries={Err, Ma-F, Mi-F},
    legend style={/tikz/every even column/.append style={column sep=0.13cm}},
    legend to name=bb,
    ]
    \addplot [red, mark=*] table [x index=0, y index=4] {data/yelp_lambda.txt};
    \addplot [blue, dashed, mark=*] table [x index=0, y index=5] {data/yelp_lambda.txt};
    \addplot [green, dashed, mark=*] table [x index=0, y index=6] {data/yelp_lambda.txt};
    \end{axis}
\end{tikzpicture}
\caption{Yelp}
\label{fig:yelp_lambda}
\end{subfigure}%
\begin{subfigure}[t]{0.32\textwidth}
  \pgfplotsset{width=\textwidth}
  \ref{bb} \\
\begin{tikzpicture}
    \begin{axis}[
    height=3.5cm,
    xlabel={$\lambda$},
    xtick={1,5,10,20,30,40,50},
  ylabel={Value},
  xmin=0, xmax=50,
        ymin=0.17, ymax=0.8,
    mark size=0.5pt,
    ymajorgrids=true,
    grid style=dashed,
    legend columns=-1,
    legend style={/tikz/every even column/.append style={column sep=0.13cm}},
    legend to name=bb,
    ]
    \addplot [red, mark=*] table [x index=0, y index=4] {data/wos_lambda.txt};
    \addplot [blue, dashed, mark=*] table [x index=0, y index=5] {data/wos_lambda.txt};
    \addplot [green, dashed, mark=*] table [x index=0, y index=6] {data/wos_lambda.txt};
    \end{axis}
\end{tikzpicture}
\caption{WOS}
\label{fig:wos_lambda}
\end{subfigure}
\begin{subfigure}[t]{0.32\textwidth}
  \pgfplotsset{width=\textwidth}
  \ref{bb} \\
\begin{tikzpicture}
    \begin{axis}[
    height=3.5cm,
    xlabel={$\lambda$},
    xtick={1,5,10,20,30,40,50},
  ylabel={Value},
  xmin=0, xmax=50,
        ymin=0.05, ymax=1.15,
    mark size=0.5pt,
    ymajorgrids=true,
    grid style=dashed,
    legend columns=-1,
    legend style={/tikz/every even column/.append style={column sep=0.13cm}},
    legend to name=bb,
    ]
    \addplot [red, mark=*] table [x index=0, y index=4] {data/qba_lambda.txt};
    \addplot [blue, dashed, mark=*] table [x index=0, y index=5] {data/qba_lambda.txt};
    \addplot [green, dashed, mark=*] table [x index=0, y index=6] {data/qba_lambda.txt};
    \end{axis}
\end{tikzpicture}
\caption{QCD}
\label{fig:qcd_lambda}
\end{subfigure}
\caption{Influence of $\lambda$ on RLHR approach with BERT.
        Err, Ma-F and Mi-F denote logical error rate, Macro-F1 and Micro-F1 respectively.}
\label{fig:lambda_fig}
\vspace{-5pt}
\end{figure*}

\subsubsection{Analyses on Rollback Algorithm}

Due to the limit of space, we only report the results of our proposed rollback algorithm based on BERT and put the results on DistilBERT in Appendix \ref{sec:distilbert_res}.
As shown in Table \ref{tab:rollback_res}, we can see that when being combined with our proposed rollback algorithm, the performance of BERT+RLHR can be further improved, raising Example-based F1 on Yelp, WOS, and QCD from 49.52\%, 64.43\%, 39.99\% to 50.01\%, 69.32\% and 40.13\%, respectively. 
Our proposed rollback algorithm can also be combined with BERT only, while the gain is relatively marginal. 
We further investigate this and observe that at the same level of the label hierarchy, the matching scores obtained in RLHR is more polarized, compared to those obtained with BERT, suggesting RLHR is more confident about the predictions when the label hierarchy is provided. This yields a better prediction performance of RLHR when the rollback algorithm is adopted. 


Meanwhile, we compare the hierarchical inference method \cite{huang2019hierarchical} with our rollback algorithm. 
Both methods can completely remove logical errors from the predicted results.
However, as we can see in the table, the performance of the hierarchical inference method is not consistent on the three datasets, with either BERT or BERT+RLHR.
When conducting hierarchical inference, BERT+RLHR achieves the best Micro-F1 and Example-based F1 on QCD dataset, while the performance is harmed with a significant gap on the WOS dataset.
Similarly, the performance of hierarchical inference with BERT achieves minor improvement on the QCD dataset, while on WOS and Yelp, the performance is sometimes improved marginally or sometimes worse.
The effectiveness of hierarchical inference method depends mainly on the classification difficulty of labels at the higher levels of label hierarchies.
As we know, such labels are usually more abstract and general, thus making the performance of hierarchical inference susceptible.


\subsubsection{Influence of $\lambda$}
\label{sec:influence_of_lambda}

We discuss the influence of the parameter $\lambda$ on logical error rates and useen label classification in this section.
Due to the limit of space, we only represent the results with BERT and put the results based on DistilBERT in Appendix \ref{sec:distilbert_lambda}.
As shown in Figure \ref{fig:lambda_fig}, for datasets with large hierarchy, like Yelp and QCD, a larger $\lambda$ helps achieve better classification performance on unseen labels, while it will bring more logical errors.
On the contrary, a relatively small $\lambda$ yields better classification performance and lower logical error rates on datasets with small hierarchies like WOS, as shown in Figure \ref{fig:wos_lambda}.
The reason is that for a large hierarchy, the number of sampled correct deduction paths will be much less than that of the wrong paths which is common in the ZS-MTC task because the positive labels are usually much less than negative labels, while for a small label hierarchy, the number of sampled correct paths are close to the false ones.
A large $\lambda$ will encourage a model to focus more on sampled correct paths, which will hence improve the classification performance.
Meanwhile, if $\lambda$ is too large, it will bring a bias to the dominating labels which appear more in the datasets. 
Thus it will reduce the generalization ability of the model, which will harm the performance.

\section{Conclusion}

We propose a Reinforced Label Hierarchy Reasoning approach to incorporate label hierarchies into pretrained models in order to better solve the zero-shot multi-label text classification tasks.
We train an agent that starts from the root label, navigates to potential labels in the label hierarchies and generates multiple deduction paths.
By rewarding based on the sampled deduction paths, our approach can strengthen the interconnections among the labels during the training stage.
To overcome the weakness of hierarchical inference methods, we further design a rollback algorithm that can remove the logical errors in flat predictions.
Experiments on the three datasets demonstrate that our proposed approach improves the performance of pretrained models and enable the models to make more consistent predictions.


\clearpage
\bibliography{naacl2021}
\bibliographystyle{acl_natbib}

\clearpage
\appendix

\section{Appendix}
\label{sec:appendix}

\subsection{Dataset preparation}
\label{sec:data_example}

We split the labels in the label space as seen labels and unseen labels.
Unseen labels do not necessarily need to be leaf labels, and if an intermediate label is chosen as unseen, then all its descendant labels will be set as unseen.
Meanwhile, each data instance in dev/test sets will contain at least one unseen label.

Table \ref{tab:dataset_example} shows the example instances of Yelp, WOS and QCD datasets used in this work.

\subsection{Rollback Results with DistilBERT}
\label{sec:distilbert_res}
As shown in Table \ref{tab:distilbert_rollback_res}, DistilBERT+RLHR with Rollback algorithm can achieve the best performance on most evaluation metrics.
Although the hierarchical inference method can improve DistilBERT on QCD dataset, its performance is not consistent. 
It lowers the performance by large margins on WOS with both DistilBERT and DistilBERT+RLHR.
In contrast, the rollback algorithm has consistent performance on all the three datasets, especially when combined with our proposed RLHR approach.

\subsection{Influence of $\lambda$ with DistilBERT}
\label{sec:distilbert_lambda}

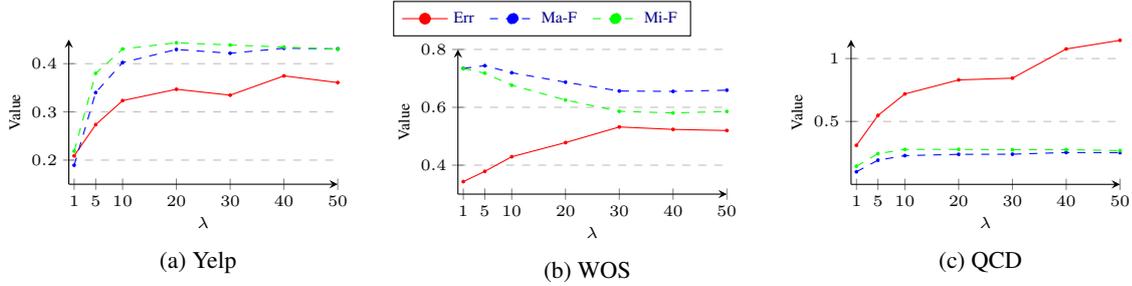
\begin{figure*}[t] \tiny 
  \begin{subfigure}[t]{0.32\textwidth}
  \pgfplotsset{width=\textwidth}
  \ref{bb} \\
\begin{tikzpicture}
    \begin{axis}[
    height=3.5cm,
    xlabel={$\lambda$},
    xtick={1,5,10,20,30,40,50},
  ylabel={Value},
  xmin=0, xmax=50,
        ymin=0.15, ymax=0.45,
    mark size=0.5pt,
    ymajorgrids=true,
    grid style=dashed,
    legend columns=-1,
    legend entries={Err, Ma-F, Mi-F},
    legend style={/tikz/every even column/.append style={column sep=0.13cm}},
    legend to name=bb,
    ]
    \addplot [red, mark=*] table [x index=0, y index=1] {data/yelp_lambda.txt};
    \addplot [blue, dashed, mark=*] table [x index=0, y index=2] {data/yelp_lambda.txt};
    \addplot [green, dashed, mark=*] table [x index=0, y index=3] {data/yelp_lambda.txt};
    \end{axis}
\end{tikzpicture}
\caption{Yelp}
\label{fig:yelp_lambda_distilbert}
\end{subfigure}%
\begin{subfigure}[t]{0.32\textwidth}
  \pgfplotsset{width=\textwidth}
  \ref{bb} \\
\begin{tikzpicture}
    \begin{axis}[
    height=3.5cm,
    xlabel={$\lambda$},
    xtick={1,5,10,20,30,40,50},
  ylabel={Value},
  xmin=0, xmax=50,
        ymin=0.3, ymax=0.8,
    mark size=0.5pt,
    ymajorgrids=true,
    grid style=dashed,
    legend columns=-1,
    legend style={/tikz/every even column/.append style={column sep=0.13cm}},
    legend to name=bb,
    ]
    \addplot [red, mark=*] table [x index=0, y index=1] {data/wos_lambda.txt};
    \addplot [blue, dashed, mark=*] table [x index=0, y index=2] {data/wos_lambda.txt};
    \addplot [green, dashed, mark=*] table [x index=0, y index=3] {data/wos_lambda.txt};
    \end{axis}
\end{tikzpicture}
\caption{WOS}
\label{fig:wos_lambda_distilbert}
\end{subfigure}
\begin{subfigure}[t]{0.32\textwidth}
  \pgfplotsset{width=\textwidth}
  \ref{bb} \\
\begin{tikzpicture}
    \begin{axis}[
    height=3.5cm,
    xlabel={$\lambda$},
    xtick={1,5,10,20,30,40,50},
  ylabel={Value},
  xmin=0, xmax=50,
        ymin=0.0, ymax=1.15,
    mark size=0.5pt,
    ymajorgrids=true,
    grid style=dashed,
    legend columns=-1,
    legend style={/tikz/every even column/.append style={column sep=0.13cm}},
    legend to name=bb,
    ]
    \addplot [red, mark=*] table [x index=0, y index=1] {data/qba_lambda.txt};
    \addplot [blue, dashed, mark=*] table [x index=0, y index=2] {data/qba_lambda.txt};
    \addplot [green, dashed, mark=*] table [x index=0, y index=3] {data/qba_lambda.txt};
    \end{axis}
\end{tikzpicture}
\caption{QCD}
\label{fig:qcd_lambda_distilbert}
\end{subfigure}
\caption{Influence of $\lambda$ on RLHR approach with DistilBERT.
        Err, Ma-F and Mi-F denote logical error rate, Macro-F1 and Micro-F1 respectively.}
\label{fig:lambda_fig_distilbert}
\end{figure*}

As shown in Figure \ref{fig:lambda_fig_distilbert}, the influence of parameter $\lambda$ on three datasets with DistilBERT is similar to that with BERT.   
For Yelp and QCD datasets, a larger $\lambda$ helps achieve better classification performance on unseen labels, while it will bring more logical errors.
On the contrary, a relatively small $\lambda$ yields both better classification performance and lower logical error rates on WOS dataset, as shown in Figure \ref{fig:wos_lambda_distilbert}.
The results support our analyses in Section \ref{sec:influence_of_lambda}.

\subsection{Deduction Path Analysis}

We represent the results of deduction paths in this section, which is an important evaluation of if the model captures the interdependencies of labels.
A path is considered as correct when it equals to or belongs to a golden deduction path, and we report Example-based Precision, Recall and F1 based on BERT.
As shown in Table \ref{tab:deduction_path_res}, BERT can achieve high recall but low precision on the deduction paths, which means that it tends to predict more labels as correct.
This is because pretrained models only take the literal tokens of labels as input without any label structure information.
On the contrary, RLHR, which incorporates the label hierarchy, can provide more accurate predictions of deduction paths with much higher precision on all the three datasets.

\begin{table}[t]
\centering
\setlength{\tabcolsep}{3.0pt}
\begin{tabular}{c|ccc|ccc}  
\toprule
    \multirow{2}{*}{Dataset} & \multicolumn{3}{c|}{BERT} & \multicolumn{3}{c}{BERT+RLHR} \\
    \cline{2-7}
            & P & R & F1 & P & R & F1 \\
    \midrule
    Yelp    & 17.17   & 72.54   & 26.03   & 38.04   & 52.61   & 40.27 \\
    WOS     & 33.25   & 77.57   & 44.35   & 47.34   & 66.51   & 53.28 \\
    QCD     & 18.43   & 58.37   & 26.68   & 22.55   & 57.11   & 30.71 \\
\bottomrule
\end{tabular}
\caption{Performance on deduction paths.
        P, R, F1 denote Example-based Precision, Recall and F1.}
\label{tab:deduction_path_res}
\end{table}

\newcommand{\tabincell}[2]{\begin{tabular}{@{}#1@{}}#2\end{tabular}}
\begin{table*}
	\begin{center}\small
		\begin{tabular}{p{2cm} p{10cm} p{2cm}}
			\toprule
			Dataset &  Text & Labels \\
			\midrule
			Yelp & 
			\tabincell{p{10cm}}{Mini donuts  at it's finest. I was there on Saturday and it was absolutely delicious. I had a mini six pack of D O's. I would highly recommend this place for a sweet snack. Five thumbs up.} & 
			\tabincell{p{2cm}}{\textit{Food},\\ \textit{Restaurants},\\ \textit{Donuts},\\ \textit{Food Stands}}\\
			\midrule
			WOS & 
			\tabincell{p{10cm}}{This paper presents the design and experimental evaluation of discrete time sliding mode controller using multirate output feedback to minimize structural vibration of a cantilever beam using shape memory alloy wires as control actuators and piezoceramics as sensor and disturbance actuator. Linear dynamic models of the smart cantilever beam are obtained using online recursive least square parameter estimation. A digital control system that consists of Simulink (TM) modeling software and dSPACE DS1104 controller board is used for identification and control. The effectiveness of the controller is shown through simulation and experimentation by exciting the structure at resonance.} & 
			\tabincell{p{2cm}}{\textit{ECE},\\ \textit{Digital control}}\\
			\midrule
			QCD &
			\tabincell{p{10cm}}{ipad usb c hub} & 
			\tabincell{p{2cm}}{\textit{Electronics},\\ \textit{Accessories \& Supplies},\\ \textit{Audio \& Video Accessories}}\\
			\bottomrule
		\end{tabular}
		\caption{Examples of the three datasets}
		\label{tab:dataset_example} 
	\end{center}
\end{table*}

\begin{table*}[t]
\centering
\setlength{\tabcolsep}{3.0pt}
\begin{tabular}{c|c|ccc|ccc|ccc}  
\toprule
    \multirow{2}{*}{Method} & \multirow{2}{*}{Setting} & \multicolumn{3}{c|}{Yelp} & \multicolumn{3}{c|}{WOS} & \multicolumn{3}{c}{QCD} \\
    \cline{3-11}
    &   &  Ma-F & Mi-F & EBF & Ma-F & Mi-F & EBF & Ma-F & Mi-F & EBF \\
    \midrule
    \multirow{2}{*}{DistilBERT} & ZS  & 
    41.42 & 40.33 & \multirow{2}{*}{30.44} & 
    70.69 & 65.19 & \multirow{2}{*}{55.18} & 
    23.68 & 24.95 & \multirow{2}{*}{33.57} \\
        & GZS & 
    21.29 & 28.18 &                      & 
    68.03 & 63.64 &                      & 
    24.43 & 34.29 &                      \\
    \midrule
    DistilBERT & ZS  & 
    41.88 & 41.00 & \multirow{2}{*}{30.61} & 
    67.81 & 66.45 & \multirow{2}{*}{53.13} & 
    21.13 & 29.29 & \multirow{2}{*}{34.35} \\
    +Hie-Infe & GZS & 
    21.49 & 28.36 &                      & 
    65.65 & 64.05 &                      & 
    23.91 & 35.12 &                     \\
    \midrule
    DistilBERT & ZS  & 
    41.49 & 40.32 & \multirow{2}{*}{30.47} & 
    70.69 & 65.19 & \multirow{2}{*}{56.54} & 
    23.81 & 24.7 & \multirow{2}{*}{33.34} \\
    +Rollback  & GZS & 
    21.28 & 28.18 &                      & 
    68.44 & 63.31 &                      & 
    24.36 & 33.99 &                     \\
    \hline\hline
    \multirow{2}{*}{DistilBERT+RLHR} & ZS  & 
    42.16 & 43.87 & \multirow{2}{*}{40.85} & 
    74.56 & 72.44 & \multirow{2}{*}{61.06} & 
    24.58 & 27.79 & \multirow{2}{*}{37.46}\\
        & GZS & 
    26.95 & 40.43 &                      & 
    71.65 & 68.05 &                      & 
    26.10 & 38.73 &                      \\
    \midrule
    DistilBERT+RLHR & ZS  & 
    39.48 & 41.65 & \multirow{2}{*}{40.65} & 
    63.61 & 64.21 & \multirow{2}{*}{53.39} & 
    20.18 & \textbf{29.68} & \multirow{2}{*}{\textbf{38.13}} \\
    +Hie-Infe  & GZS & 
    26.79 & 40.44 &                      & 
    62.63 & 64.05 &                      & 
    24.98 & \textbf{39.44} &                     \\
    \midrule
    DistilBERT+RLHR & ZS  & 
    \textbf{42.27} & \textbf{43.91} & \multirow{2}{*}{\textbf{41.03}} & 
    \textbf{74.56} & \textbf{72.44} & \multirow{2}{*}{\textbf{65.64}} & 
    \textbf{24.89} & 28.34 & \multirow{2}{*}{37.45} \\
    +Rollback & GZS & 
    \textbf{26.97} & \textbf{40.55} &                      & 
    \textbf{73.14} & \textbf{71.48} &                      & 
    \textbf{26.17} & 38.68 &                     \\
\bottomrule
\end{tabular}
\caption{Results and comparisons of our matching-score-based rollback algorithm on DistilBERT.
        Ma-F, Mi-F, EBF, Err denote Macro-F1, Micro-F1, Example-based F1 and logical error rate respectively, and ZS, GZS denote zero-shot setting and generalized zero-shot setting.
        Bold figures indicate the best results for each metric.}
\label{tab:distilbert_rollback_res}
\end{table*}

\end{document}